\documentclass{article}

\usepackage{PRIMEarxiv}

\usepackage[utf8]{inputenc} 
\usepackage[T1]{fontenc}    
\usepackage{hyperref}       
\usepackage{url}            
\usepackage{booktabs}       
\usepackage{amsfonts}       
\usepackage{nicefrac}       
\usepackage{microtype}      
\usepackage{lipsum}
\usepackage{amsmath}
\usepackage{bm}
\usepackage{fancyhdr}       
\usepackage{graphicx}       
\graphicspath{{media/}}     

\title{Fine-Tuning Pre-trained Large Time Series Models for Prediction of Wind Turbine SCADA Data}

\author{
  Yuwei Fan \\
  Department of Energy and Power Engineering \\
  Tsinghua University \\
  Beijing, China \\
  \texttt{fan-yw22@mails.tsinghua.edu.cn} \\
   \And
  Tao Song \\
  Department of Energy and Power Engineering \\
  Tsinghua University \\
  Beijing, China \\
  \texttt{songtao21@mails.tsinghua.edu.cn} \\
  \AND
  Chenlong Feng \\
  Department of Energy and Power Engineering \\
  Tsinghua University \\
  Beijing, China \\
  \texttt{fengcl23@mails.tsinghua.edu.cn} \\
  \And
  Keyu Song \\
  Department of Energy and Power Engineering \\
  Tsinghua University \\
  Beijing, China \\
  \texttt{sky20@mails.tsinghua.edu.cn} \\
  \And
  Chao Liu\thanks{Corresponding author.} \\
  Department of Energy and Power Engineering \\
  Tsinghua University \\
  Beijing, China \\
  \texttt{cliu5@tsinghua.edu.cn} \\
  \And
  Dongxiang Jiang \\
  Department of Energy and Power Engineering \\
  Tsinghua University \\
  Beijing, China \\
  \texttt{jiangdx@tsinghua.edu.cn} \\
}

\begin{document}
\maketitle

\begin{abstract}
The remarkable achievements of large models in the fields of natural language processing (NLP) and computer vision (CV) have sparked interest in their application to time series forecasting within industrial contexts.  This paper explores the application of a pre-trained large time series model, Timer, which was initially trained on a wide range of time series data from multiple domains, in the prediction of Supervisory Control and Data Acquisition (SCADA) data collected from wind turbines.  The model was fine-tuned on SCADA datasets sourced from two wind farms, which exhibited differing characteristics, and its accuracy was subsequently evaluated.  Additionally, the impact of data volume was studied to evaluate the few-shot ability of the Timer.  Finally, an application study on one-turbine fine-tuning for whole-plant prediction was implemented where both few-shot and cross-turbine generalization capacity is required. The results reveal that the pre-trained large model does not consistently outperform other baseline models in terms of prediction accuracy whenever the data is abundant or not, but demonstrates superior performance in the application study.  This result underscores the distinctive advantages of the pre-trained large time series model in facilitating swift deployment.
\end{abstract}

\keywords{Large Time Series Model \and Time Series Prediction \and Wind Turbine \and Few-shot}

\section{Introduction}
\label{sec:intro}
The time series prediction of data collected by Supervisory Control and Data Acquisition (SCADA) system of wind turbines is a pivotal challenge as it serves as a foundational method of artificial intelligence (AI) in various aspects of wind turbine operations and maintenance (O\&M), including data preprocessing\cite{RN200,RN320}, fault diagnosis\cite{RN188,RN285,RN296}, and wind power forecasting\cite{RN290,RN313}. Within the sphere of data preprocessing, time series prediction is employed to fill in missing values by leveraging historical monitoring data. For fault diagnosis, the distribution of discrepancies between actual monitored values and those generated by time series predictions is examined to identify anomalies. As for the domain of wind power forecasting, the outcomes of time series predictions are directly applied. Therefore, as the cornerstone of these applications, the enhancement of time series prediction accuracy is crucial for improving their effectiveness and reliability, which in turn contributes to the efficiency and profitability of renewable energy generation.

The existing time series prediction methods are primarily based on deep learning, including recurrent neural network (RNN)\cite{RN334}, graph neural networks (GNN)\cite{RN335}, and the transformer\cite{RN206}. Deep learning methods have made advancements in precision compared to classical forecast methods, but they have not yet demonstrated the advantages brought by the increase in parameter scale of large models, which has been conducted in the fields of NLP and CV\cite{RN336,RN337,RN216,RN211}. The powerful problem-solving capabilities and the “one for many” generalization performance demonstrated by large models are impressive, and these abilities can play a significant role in the O\&M of wind turbines. The wind turbine O\&M is constantly challenged by the need for enhanced precision. Additionally, it necessitates a versatile approach capable of generalizing across various turbine and facilities to accommodate the diversity in data patterns, thereby streamlining the effort involved in developing and sustaining models.

There are two methods for applying large models to time series prediction: adjusting the large language models to fit the data patterns of time series, and training foundation models from scratch on time series data. The approaches to adapt large language models to time series include two types: text-visible LLM adaption and embedding-visible LLM adaption\cite{RN327}. The former\cite{RN291} converts numerical time series data into strings and integrates them into prompts along with other contextual information. The prompts are then processed by LLM which generates outputs based on the inference capabilities. The latter\cite{RN292} embeds numerical time series data into vector sequences and fine-tunes the embedding layer on downstream time series datasets to align the vector representations of time series with those of text. These two methods require fewer computational resources compared to training a large model from scratch and are more easily integrated with text for multimodal analysis. However, the lack of evidence that time series data exhibit patterns similar to language raises doubts about the appropriateness of using large language models for time series forecasting, and existing literature has highlighted the limitations of these approaches, necessitating the development of large time series models specifically trained on time series data\cite{RN328}.

Training a foundation model for time series from scratch usually involves preprocessing time series data and designing model architecture. In the data preprocessing stage, data quality management is necessary, and due to the limited amount of time series data in a single domain, alignment of the data formats across multiple domains is required. After data preparation, large-scale pre-training is typically conducted based on the transformer architecture. As the training data encompasses time series from diverse domains, the model acquires a broad-spectrum capability for analyzing time series data across a variety of domains.

To ascertain the applicability of the large time series foundation model within specialized domains and to delineate its comparative strengths and weaknesses against conventional models, this study employs a pre-trained large time series model named Timer\cite{RN314} for the prediction of wind turbine SCADA data. Timer encompasses 67 million parameters and has been initialized with pre-training time series datasets spanning various sectors. The experiments organized and preprocessed the SCADA data from two real-world wind farms, which were subsequently used for fine-tuning and evaluation of the Timer. The variety in data volume and plant type (onshore or offshore) enables a comprehensive evaluation of the large time series model’s performance across varying conditions. For the wind farm with adequate data, the fraction of data employed in the fine-tuning was modified to further scrutinize the influence of data volume on the model’s predictive accuracy. This work also designed an application scenario of one-turbine fine-tuning for whole-plant prediction, which necessitates both the few-shot and generalization capabilities of the model. Experimental results across different wind farms and data volumes indicate that large time series model does not exhibit a dominant accuracy advantage in data-sufficient conditions. In scenarios with limited data, pre-trained large time series models demonstrate few-shot learning advantages, although this capability does not significantly surpass that of LSTM in short-term predictions. In the context of fine-tuning on a single turbine, the large model not only exhibits few-shot learning ability but also generalization across turbines, yielding comprehensive accuracy advantages over other methods in this scenario. This demonstrates the strengths of pre-trained large time series models for rapid deployment in wind farms.

The contributions of this paper include: (1) This paper applies the large time series foundation model to the time series prediction of real-world SCADA data; (2) A comparative study was conducted to evaluate the strengths and weaknesses of large time series model against conventional methods across varying data volumes and prediction horizons; (3) The few-shot learning and generalization benefits of the large time series model was illustrated within the context of single-turbine fine-tuning for whole-plant prediction, offering valuable insight for future application of large time series models in wind turbine SCADA data.

The remaining sections are outlined as follows. Section \ref{sec:method} presents the methodology, including the data preprocessing and the architecture of the model; Section \ref{sec:exp} describes the implementation of experiments based on three real-world SCADA dataset, along with the presentation of results. Finally, the conclusions are given in Section \ref{sec:conc}.

\section{Methodology}
\label{sec:method}

\subsection{Timer}
The Timer was proposed in reference\cite{RN314}, which is a 67-million-parameter transformer model that has undergone large-scale pre-training on time series datasets across multiple fields. It is pre-trained by performing autoregressive next-token predictions on the time series, demonstrating promising performance in various downstream tasks across multiple domains.

\subsubsection{Data for pre-training}
The Timer model was pre-trained on the Unified Time Series Dataset (UTSD), a high-quality collection of time series data spanning various domains. The dataset integrates 29 distinct datasets, classified into ten different fields according to their origin, such as energy, environment, health, and others. This dataset incorporates time series with diverse sampling rates, extending from yearly to minute-by-minute intervals. UTSD implements a series of strategies to guarantee the quality of the data. Initially, missing values in the data are addressed through linear interpolation; subsequently, comprehensive statistical analyses are performed to evaluate stationarity and forecastability. These metrics facilitate the identification of four superior subsets within UTSD, each escalating in complexity and variety of patterns in relation to their size. The most extensive subset, UTSD-12g, encompasses one billion data points. The vast scale of UTSD ensures comprehensive model training and the acquisition of generalized time series patterns across a multitude of domains.

\subsubsection{Single-Series Sequence}
\begin{figure}
  \centering
  \includegraphics[width=0.8\linewidth]{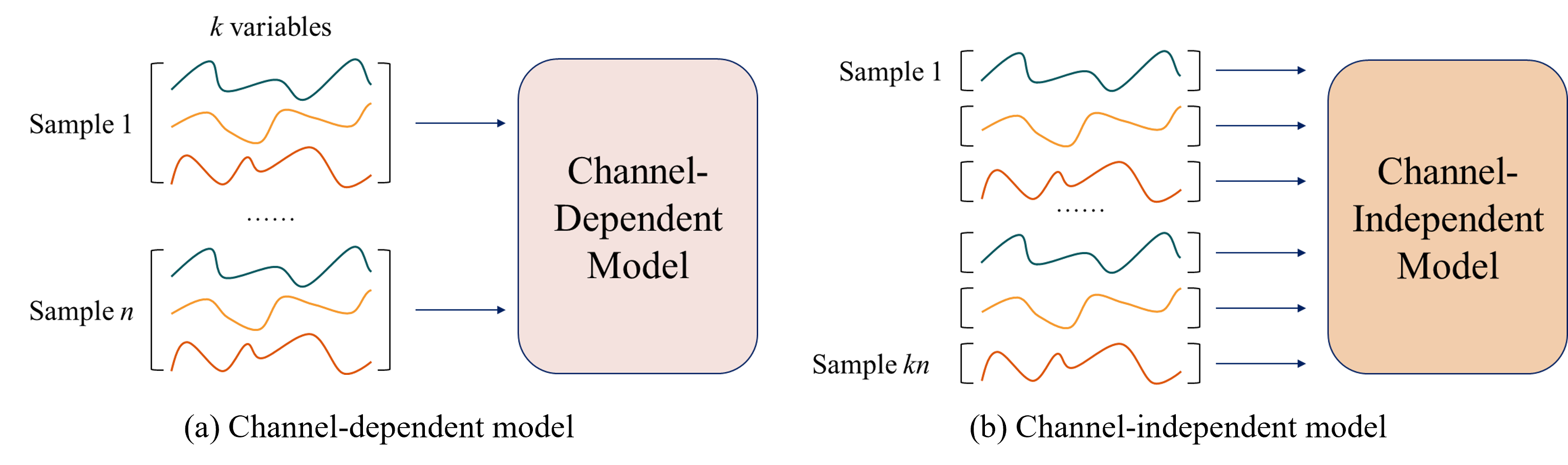}
  \caption{The difference between channel-dependent and channel-independent model.}
  \label{fig:fig1}
\end{figure}

The Timer employs the single-series sequence (S3) format to standardize heterogeneous data from disparate sources, accommodating the formatting discrepancies that arise due to variations in the number of variables, sampling frequencies, and other attributes. S3 is an iteration of the channel-independent (CI) strategy. CI splits multivariate time series into univariate series, with each variable analyzed independently, thereby disregarding the inter-variable correlations. In contrast, conventional time series modeling often utilizes a channel-dependent (CD) strategy, which entails simultaneous modeling of multiple variables, taking into account their inter-dependencies. The difference between these two strategies is delineated in Figure \ref{fig:fig1}. Although the CI approach neglects the important correlations between variables, it offers a convenient means of harmonizing such diverse data. Additionally, researches\cite{RN317,RN342} have shown that the CI, compared to CD strategy, offers better adaptability, lower training data requirements, and reduced risk of overfitting, leading to higher precision in time series modeling.

S3 extends the CI approach by not requiring temporal alignment within the same batch of time series samples, nor do the samples need to originate from the same dataset. Additionally, S3 implements instance normalization, ensuring that the training time points for each univariate sequence adhere to a normal distribution. This process helps to neutralize disparities in amplitude across various datasets.

\subsubsection{Tokenization}
The Timer adopts a strategy of segmenting univariate time series into patches for the purpose of tokenization. Conventionally, each time point within a series is treated as an individual token. However, due to the quadratic relationship between the computational complexity of attention mechanisms and sequence length, tokenizing by individual points results in a very high computational complexity for long sequences. Furthermore, the information of a single time point often pales in comparison to the importance of local trend characteristics. As a result, the practice of dividing time series into patches of a defined size as tokens has emerged as a prevalent technique. The Timer utilizes non-overlapping windows for segmentation and the tokenization formula is presented as follows:
\begin{equation}
    \bm{s}_i=\{X_{(i-1)S+1},…,X_{iS}\} \in \mathbb{R}^S,i=1,…,N
\end{equation}
where $\bm{X}=\{X_{1},…,X_{NS}\} \in \mathbb{R}^{NS}$ is the original time series, and $\bm{s}={\bm{s}_1,…,\bm{s}_N} \in \mathbb{R}^{N \times S}$ is the token sequence.

\subsubsection{Decoder-Only Transformer}
\begin{figure}
    \centering
    \includegraphics[width=0.8\linewidth]{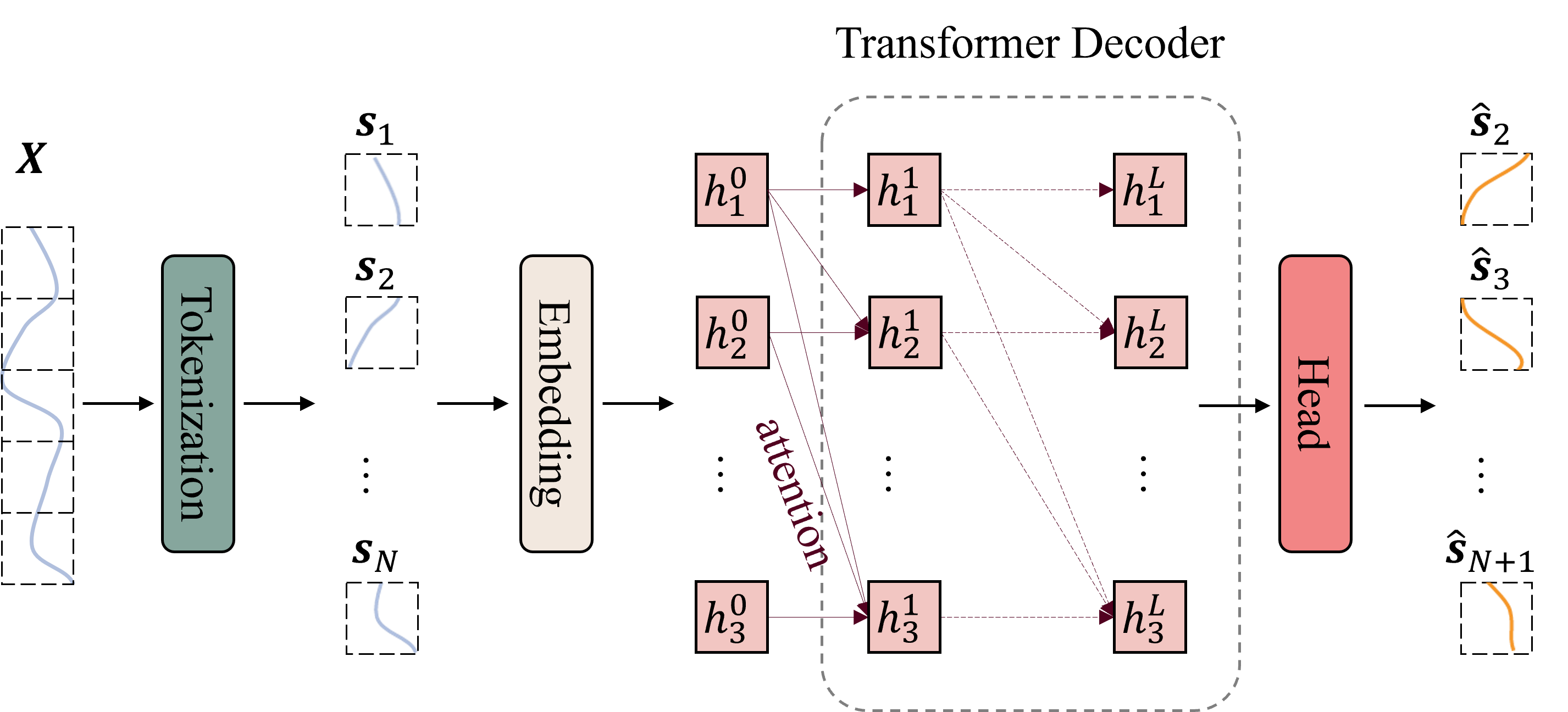}
    \caption{The architecture of Timer.}
    \label{fig:fig2}
\end{figure}

The Timer leverages a decoder-only transformer architecture for time series modeling. The distinction between encoder and decoder modules has been a fundamental aspect of transformer design since its inception, with the encoder facilitating global bidirectional attention for the extraction of features from time series data, and the decoder utilizing an attention mask to enforce unidirectional attention, thereby precluding tokens from attending to subsequent positions in the sequence. Inspired by generative pre-training (GPT), the decoder-only transformer configuration has increasingly become the norm, as researches\cite{RN343,RN344} suggest that models based on this architecture demonstrate enhanced generalization capabilities. In alignment with this trend, the Timer model adopts the decoder-only transformer block for its time series modeling. The model architecture of the Timer is depicted in Figure \ref{fig:fig2}, and the equation of the model is presented as follows:
\begin{equation}
    \begin{aligned}
        \bm{h}^0_i&=\bm{s}_i\bm{W}_e + \bm{TE}_i, i=1,...,N \\
        \bm{H}^l&={\rm TrmBlock}(\bm{H}^{l-1}), l=1,...,L \\
        \{\bm{s}_{i+1}\}&= \bm{H}^L\bm{W}_d, i = 1,...,N
    \end{aligned}
\end{equation}
where $\bm{W}$ represents weights, $\bm{TE}$ stands for the optional time stamp embedding, and $\bm{H}=\{\bm{h}_i\}$ is the latent representation of the embedding and transformer blocks.

Similar to GPT, the Timer model also utilizes autoregression as its pre-training task, a self-supervised learning method that equips the model with the capability to predict next token in a time series. This approach facilitates the model’s ability to learn from extensive, unlabeled time series datasets. The objective function for its pre-training task is presented as follows:
\begin{equation}
    \mathcal{L}_{MSE}=\frac{1}{NS}\sum\mid\mid\bm{s}_i-\hat{\bm{s}}_i\mid\mid^2_2,i=2,...,N
\end{equation}

\subsection{Fine-tune on SCADA dataset}
\subsubsection{Data Preprocessing}
Prior to fine-tuning the pre-trained large time series model, it is essential to preprocess the SCADA data obtained from real-world wind farms to guarantee data quality. To exclude the influence of wind curtailment, sensor noise, transient errors, and equipment failures, this work implements a data cleaning strategy designed to eliminate outliers by leveraging key variables within the wind turbine SCADA dataset.

\begin{figure}
    \centering
    \includegraphics[width=0.8\linewidth]{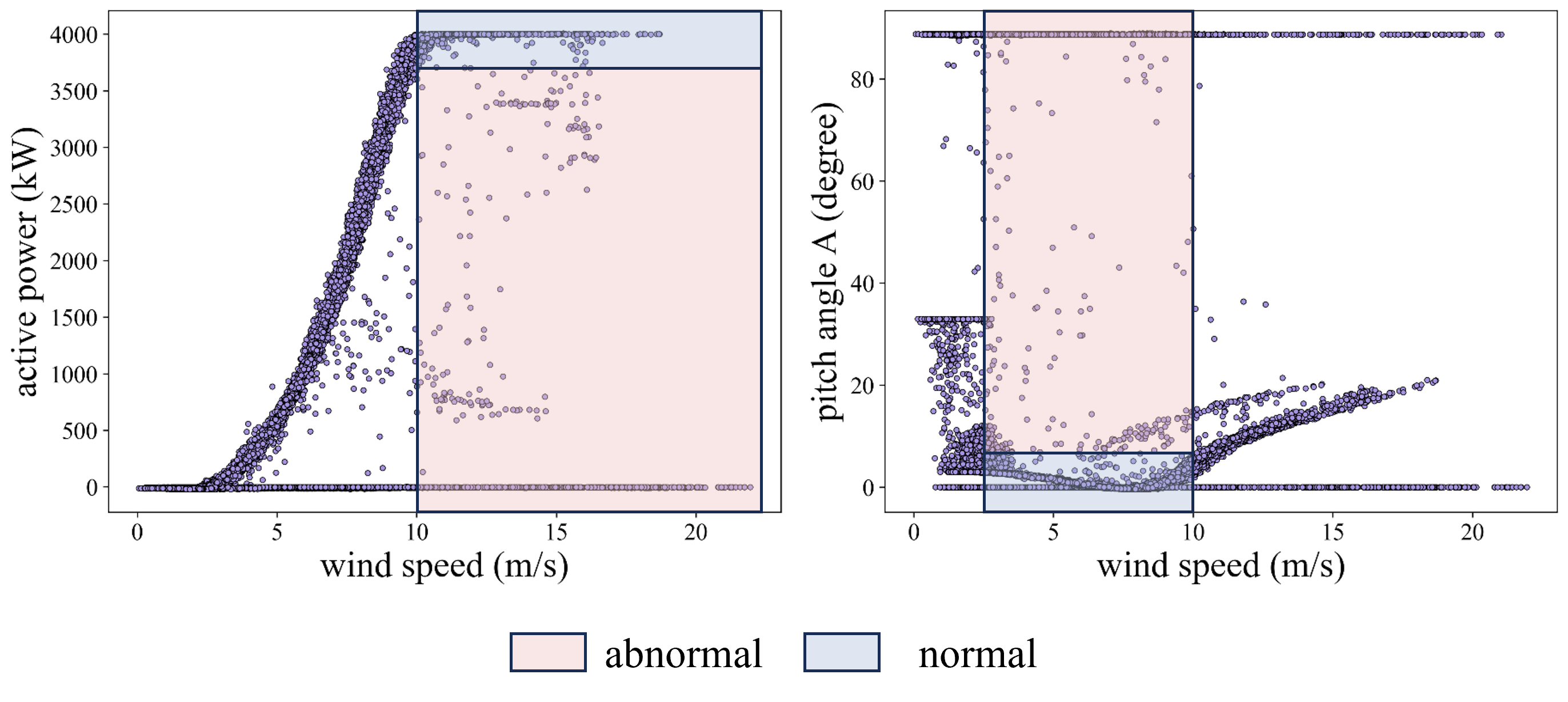}
    \caption{The principle of the data cleaning method.}
    \label{fig:fig3}
\end{figure}

Initially, an examination of the distribution of each monitoring parameter within the SCADA data was conducted, removing outliers likely resulting from sensor damage or data transmission issues. Subsequently, with consideration to the operational principles of wind turbines, additional data cleansing was performed by referencing the turbine’s working states across different wind speed ranges. Below the cut-in wind speed, the turbine is disconnected from the grid, and the power output is nil. In the range between the cut-in wind speed and the rated wind speed, which is the maximum power point tracking (MPPT) segment, the turbine maintains a blade pitch angle of zero degrees, thus achieving the highest wind energy utilization coefficient, where the relationship between power and wind speed is approximately cubic. From the rated wind speed to the cut-off wind speed, the turbine incrementally increases the blade pitch angle in response to rising wind speeds and ensure a constant rated power output. Consequently, leveraging these principles, outliers primarily due to power limiting were identified and removed through the interrelation of wind speed, power, and blade pitch angle: within the MPPT segment, an upper limit for the blade pitch angle was established, with values exceeding this limit being flagged as outliers; in the range between the rated and cut-off wind speed, a lower limit for power was set, and data points falling below this threshold were excluded. The principle of this cleaning step is shown in Figure \ref{fig:fig3}. Subsequently, density-based spatial clustering of applications with noise (DBSCAN) and local outlier factor (LOF) were applied for the final refinement of the SCADA data, with the objective of excluding measurement errors attributable to sensor malfunctions. 

After data cleaning, the distribution of normal data points on the time axis becomes discontinuous. In response to this situation, this paper replaces individual existing outliers with linear interpolation and, according to the required sequence length of the time series, applies a sliding window to continuous normal time points to create time series samples, ensuring that the patterns of the time series are not disrupted.

\subsubsection{Fine-tune and Inference}
The fine-tuning of the pre-trained Timer model on SCADA datasets for downstream tasks encompasses several key steps. Initially, the processed time series samples are transformed into the S3 format. Subsequently, given the simplicity of the autoregressive pre-training task for time series prediction, the finetuning also involves the prediction of the next token, utilizing the loss function detailed in Equation (3). Throughout the fine-tuning process, the Timer backbone parameters remain unfrozen, and no additional regression head is appended to the Timer model. Instead, a small learning rate is applied uniformly across the Timer model, with the Adam optimizer employed for optimization. Following fine-tuning, in a manner consistent with models that typically utilize autoregression, inference is carried out by iteratively forecasting the next token. The number of iterations is determined by the required prediction horizon, and the requisite prediction length is subsequently extracted.

\section{Experiment}
\label{sec:exp}
\subsection{Dataset and Preprocessing}
\begin{figure}
    \centering
    \includegraphics[width=0.8\linewidth]{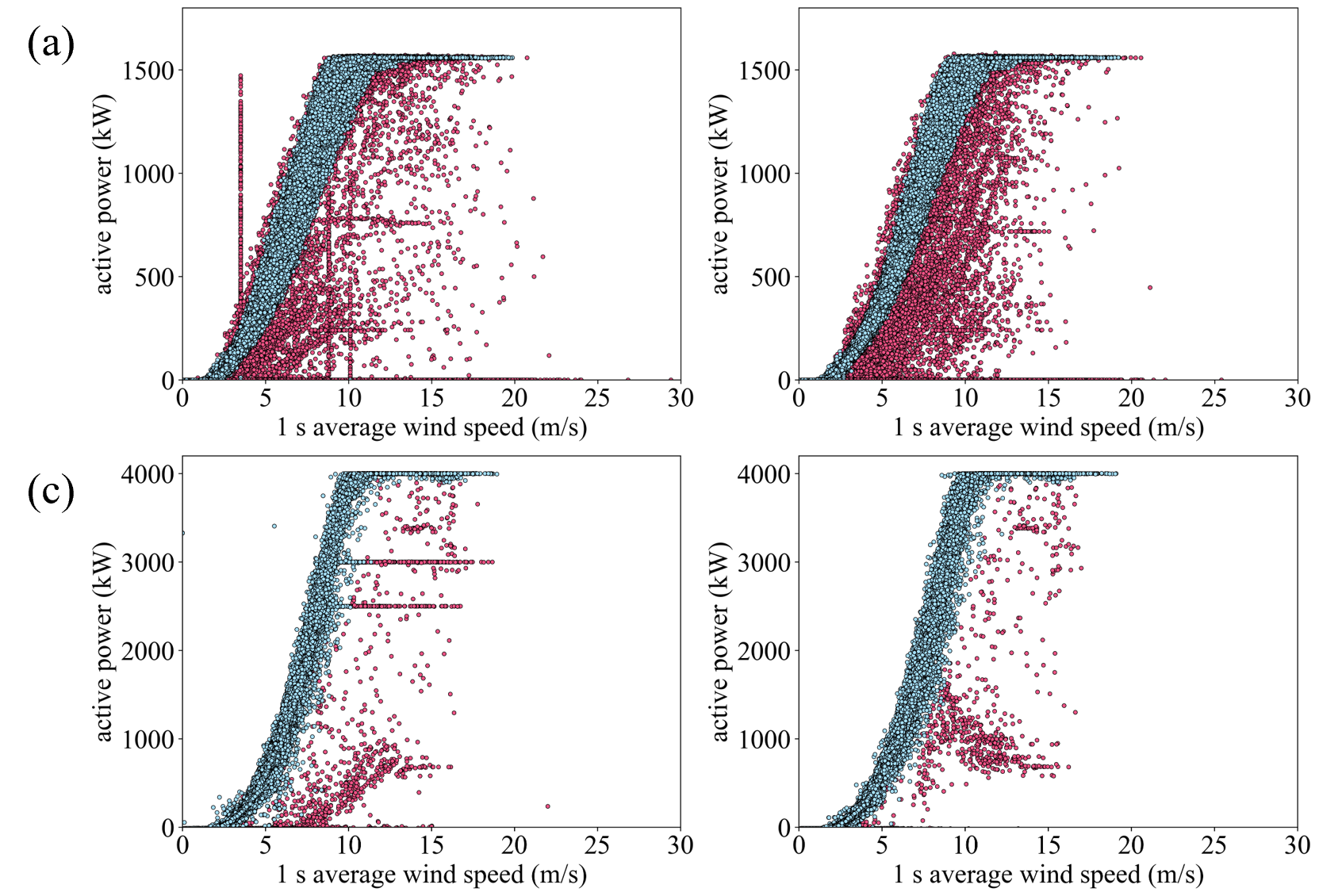}
    \caption{Results of data cleaning. (a) Plant 1. (b) Plant 2. (C) Plant 3.}
    \label{fig:fig4}
\end{figure}

In the experiment, real-world SCADA datasets obtained from two wind farms located differently in China are utilized. Plant 1, an onshore plant equipped with 64 turbines each rated at 1.5 MW, provided data spanning three years. The initial two years of data were allocated for constructing the training sets, while the final year was divided equally between validation and testing, each encompassing six months of data. Plant 2, an offshore wind farm with 121 turbines each rated at 4.0 MW, also provided a one-year dataset, where the training set comprised six months, and the validation and test sets each comprised three months. The SCADA data from all two wind farms were sampled at 10-minute intervals. The raw data underwent the cleaning process to eliminate outliers, and the efficacy of this method across the three wind farms is illustrated in Figure \ref{fig:fig4}.

Four key variables were selected from the SCADA attributes to construct multivariate time series: wind speed, power, generator speed, and ambient temperature. A sliding window was employed to generate the time series dataset from the continuous normal data points, with a window size of 768. For the training and validation sets, the window slid in increments of 100, whereas for the test set, the increment was 1. The numbers of samples used for training, validation, and testing are detailed in Table \ref{tab:tab1}. Plant 1 offers a relatively large sample size, facilitating the assessment of the performance of large time series models in data-rich environments. Conversely, plant 2 present smaller sample sizes, which are suitable for evaluating whether large time series models demonstrate distinct properties compared to traditional models in few-shot scenarios. The differences among these two wind farms in terms of wind farm type, turbine model, and data volume allow the experiment to cover a broad range of wind farm application scenarios.

\begin{table}
 \caption{Number of samples in different dataset}
  \centering
  \begin{tabular}{cccc}
    \toprule 
    Plant & Train & Validation & Test \\
    \midrule
    1 & 6654 & 570 & 118347 \\
    2 & 153 & 195 & 444 \\
    \bottomrule
  \end{tabular}
  \label{tab:tab1}
\end{table}

\subsection{Experiment Settings}
This study encompasses three experiments. In the first experiment, the Timer was fine-tuned on the datasets obtained from the two wind farms and its time series prediction accuracy was assessed on the corresponding test sets. To accommodate various time series prediction scenarios, multiple prediction lengths were employed for the testing phase. Furthermore, the experiment includes a comparison between the Timer without pre-training or fine-tuning, and three additional time series prediction models: long short-term memory networks (LSTM), two transformer models with varying numbers of parameter. This comparative analysis across different scenarios serves to validate the applicability and benefits of the large time series model for predictions. The second experiment, carried out on plant 1, changed the quantity of samples used for training to delve deeper into the performance of the Timer under varying data volumes, particularly in few-shot learning scenarios, and to ascertain its comparative advantages over traditional models.  The third experiment constitutes an application study, with the objective of utilizing the few-shot learning capabilities of large models and assessing their generalization ability. Data from a single wind turbine at Plant 1 was utilized for model fine-tuning or training, following which the model was applied to all turbines within the wind farm. The experiment is supposed to demonstrate the potential for rapid deployment inherent in large models, thereby circumventing the requirement for extensive data collection procedures. The Mean Squared Error (MSE) was used to evaluate the accuracy of models.

The declaration of the hyperparameters for several models involved in the experiments is as follows. The Timer architecture comprised 8 transformer decoder blocks, each with a model dimension of 1024, a feed-forward network (FFN) comprising 2048 hidden units, an 8-head multi-head self-attention mechanism, and a dropout rate of 0.1. Including both the input embedding layer and the output regression head, the Timer model encompassed a total of 67.40 M parameters. In the process of tokenization, a sequence of 768 time points was segmented into 8 patches, each consisting of 96 time points. The initial 7 tokens were employed as the historical context for predicting the 8th token, thereby deriving the temporal trend over a span of 96 time points. Actually, in the experimental setup, all parameters remained fixed, with the exception of the dropout rate and the number of tokens for the input, to ensure consistency with the Timer architecture as pre-trained in the literature \cite{RN314}. In the experimental setup, the Timer was examined across three distinct training paradigms: training from scratch, utilizing pre-trained weights without further fine-tuning, and fine-tuning the pre-trained model. These configurations are designated as Timer-scratch, Timer-pretrained, and Timer-finetuned, respectively. Of particular interest is the performance of the Timer-finetuned model, whereas the other two configurations are primarily used to elucidate the effects of pre-training and fine-tuning on model efficacy through comparative analysis.

The Transformer model adopted a channel-dependent and decoder-only architecture, utilizing a patch length of 96. For the model with a larger scale of parameters, the configuration of the transformer blocks aligned with that of the Timer, yielding a total parameter count that closely mirrors the Timer, at approximately 68.00 million. For the Transformer with fewer parameters, the architecture was streamlined by reducing the number of transformer blocks to 4, each with a model dimension of 256 and a FFN comprising 512 hidden units. This model is referred to as transformer-mini, and it contains a total of 2.31 million parameters. The LSTM model, on the other hand, was structured as a unidirectional network, featuring 128 hidden units, a depth of three layers, and a dropout rate of 0.1, processing the time series data sequentially step by step without patching.

All models were trained utilizing the Adam optimizer. For models trained from scratch, an initial learning rate of $1\times10^{-4}$ with a cosine decay scheduler was employed across a total of 2000 epochs. Fine-tuning was performed with a reduced learning rate of $5\times10^{-6}$ for 100 epochs. The number of epochs was chosen to ensure convergence across varying data volumes, and early stopping was implemented for both training and fine-tuning phases, based on the accuracy achieved on the validation set.

\subsection{Results}
\subsubsection{Plant 1}
The MSE of different methods across various prediction length on the initial wind farm dataset is detailed in Table \ref{tab:tab2}, with the optimal outcomes for each prediction length highlighted in bold. The results reveal that the LSTM model attains the highest precision for short prediction length, whereas the Timer-finetuned outperforms others as the prediction length extends. Moreover, when the prediction length equals 96, the Transformer-mini model demonstrates the greatest accuracy. The variation in accuracy with prediction length is depicted in Figure 5. 

\begin{table}
    \caption{MSE on Plant 1.}
    \centering
    \begin{tabular}{ccccccccc}
    \toprule
    ~ & \multicolumn{6}{c}{Prediction Length} \\
    \cmidrule(r){2-7}
    ~ & 1 & 6 & 12 & 24 & 48 & 96 \\ 
    \midrule
    Timer-finetuned & 0.0459  & 0.1083  & \textbf{0.1579}  & \textbf{0.2310}  & \textbf{0.3264}  & 0.4269   \\
    Timer-pretrained & 0.0511  & 0.1211  & 0.1846  & 0.2799  & 0.3969  & 0.5120   \\ 
    Timer-scratch & 0.0609  & 0.1192  & 0.1692  & 0.2439  & 0.3388  & 0.4217   \\ 
    Transformer & 0.1323  & 0.1725  & 0.2100  & 0.2648  & 0.3357  & 0.4128   \\ 
    Transformer-mini & 0.1391  & 0.1741  & 0.2081  & 0.2620  & 0.3310  & \textbf{0.4078}   \\ 
    LSTM & \textbf{0.0325}  & \textbf{0.1035}  & 0.1671  & 0.2710  & 0.4336  & 0.6584   \\ 
    \bottomrule
    \end{tabular}
    \label{tab:tab2}
\end{table}

\begin{figure}
    \centering
    \includegraphics[width=0.6\linewidth]{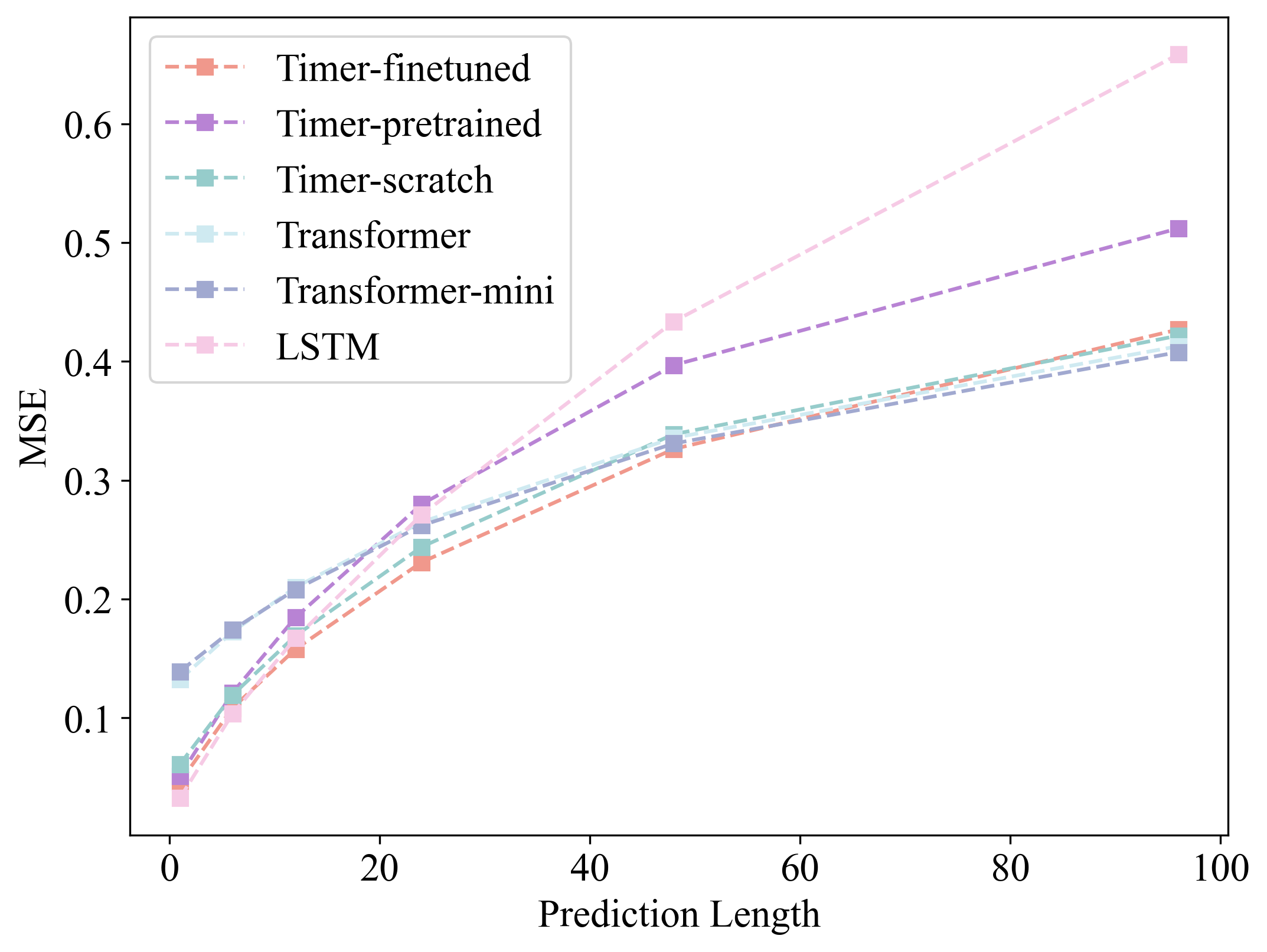}
    \caption{Variation of accuracy with prediction length in Plant 1.}
    \label{fig:fig5}
\end{figure}

The results imply that the three models are suitable for distinct predictive scenarios, and that the pre-trained and fine-tuned large models do not universally outperform across all prediction tasks. The LSTM model, with its step-wise modeling approach, provides superior accuracy for single-step predictions but is less effective for long-term forecasting due to error compounding. In contrast, models that segment data into patches for tokenization exhibit a comparative advantage in long-term predictions. Additionally, an analysis of the three Timer models illustrates the influence of pre-training and fine-tuning on model performance. As anticipated, the Timer that has undergone both pre-training and fine-tuning achieves the highest level of accuracy. The model that is pre-trained but not fine-tuned demonstrates a degree of accuracy in the downstream task of time series prediction of SCADA data. The results across varying prediction lengths indicate that pre-training notably enhances short-term prediction accuracy, while fine-tuning is more instrumental in improving long-term prediction performance. The comparison between the Transformer and Transformer-mini models indicates that an increase in parameters contributes to improved accuracy for short-term predictions, whereas it results in inferior performance for long-term predictions.

\subsubsection{Plant 2}
\begin{table}
    \caption{MSE on Plant 2.}
    \centering
    \begin{tabular}{ccccccccc}
    \toprule
    ~ & \multicolumn{6}{c}{Prediction Length} \\
    \cmidrule(r){2-7}
    ~ & 1 & 6 & 12 & 24 & 48 & 96 \\ 
    \midrule
    Timer-finetuned & \textbf{0.0268}  & 0.1496  & 0.2277  & \textbf{0.2890}  & \textbf{0.3523}  & \textbf{0.5334}   \\ 
    Timer-pretrained & 0.0320  & 0.1605  & 0.2429  & 0.2979  & 0.3593  & 0.6728   \\ 
    Timer-scratch & 0.7371  & 0.6305  & 0.6491  & 0.5657  & 0.5889  & 0.6193   \\ 
    Transformer & 0.4751  & 0.5342  & 0.5400  & 0.6006  & 0.6736  & 0.7169   \\ 
    Transformer-mini & 0.7771  & 0.5657  & 0.5631  & 0.5981  & 0.6630  & 0.6814   \\ 
    LSTM & 0.0290  & \textbf{0.1477}  & \textbf{0.2273}  & 0.3295  & 0.4702  & 0.5815   \\ 
    \bottomrule
    \end{tabular}
    \label{tab:tab3}
\end{table}

\begin{figure}
    \centering
    \includegraphics[width=0.6\linewidth]{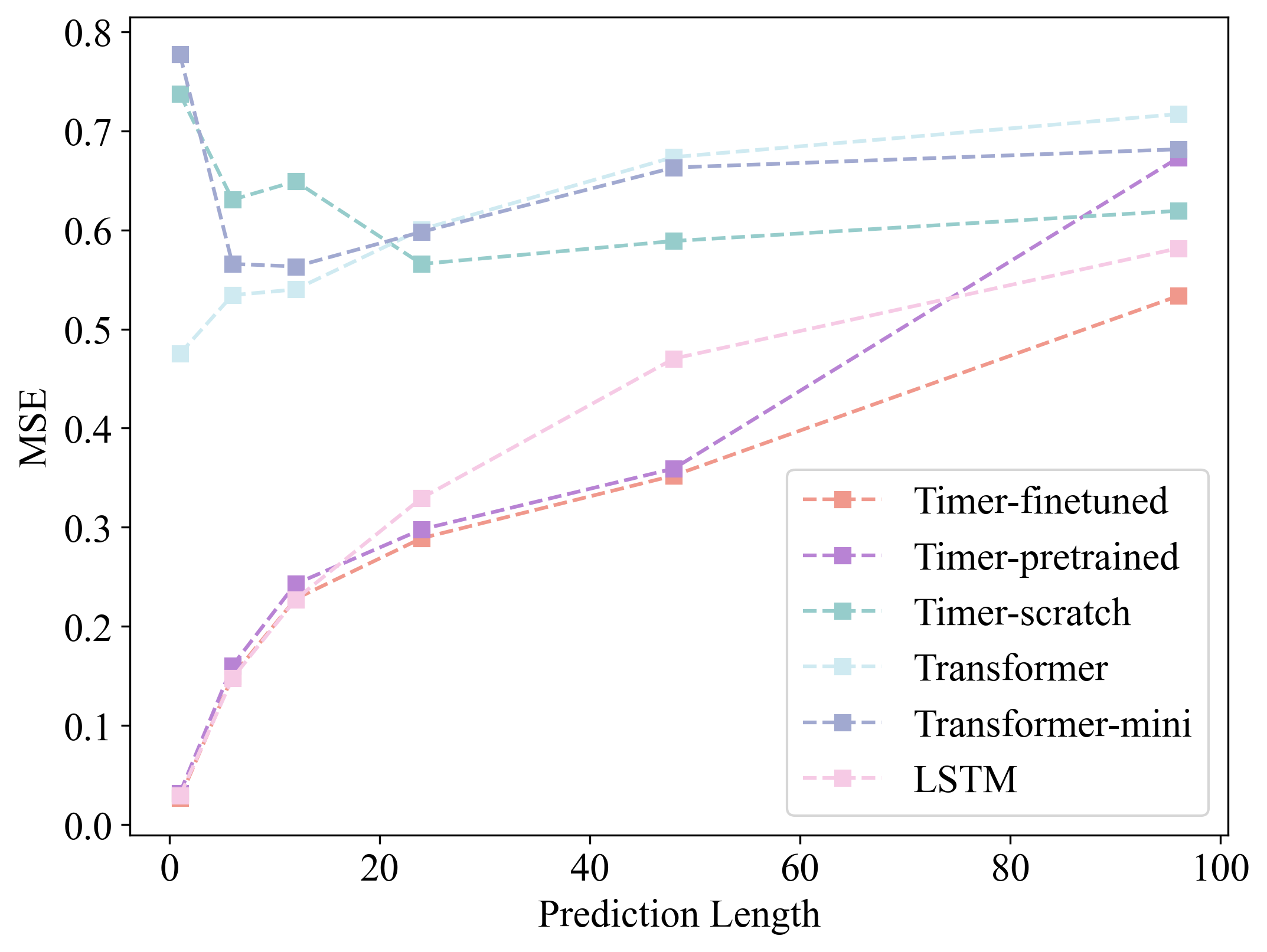}
    \caption{Variation of accuracy with prediction length in Plant 2.}
    \label{fig:fig6}
\end{figure}

On Plant 2, the MSE for different methods is presented in Table \ref{tab:tab3} and Figure \ref{fig:fig6}. The results indicate that, on this dataset with insufficient data, the Timer that has undergone pre-training and fine-tuning demonstrates a more pronounced comparative advantage over other methods featuring a large parameter scale. The enlarged accuracy gap between the pre-trained and scratch models indicates that pre-training confers enhanced few-shot capabilities to the model, allowing for better prediction performance with less data. However, this advantage is more effective in the short term. For the transformer and transformer-mini models, which have a large scale of parameters, the impact of data quantity on accuracy is significant, resulting in a loss of predictive capability for both short-term and long-term prediction. Comparison between the two models reveals that an increase in the scale of parameters may lead to a decrease in accuracy, potentially due to overfitting when faced with limited data. This finding also underscores the importance of pre-training for the application of large models. For LSTM, which has fewer parameters than the transformer-based models, the influence of data volume on predictive accuracy is relatively diminished. Short-term accuracy is comparable to that of the Timer-finetuned but continues to demonstrate suboptimal performance in long-term forecasting.

\subsubsection{The influence of data volume}
\begin{figure}
    \centering
    \includegraphics[width=0.8\linewidth]{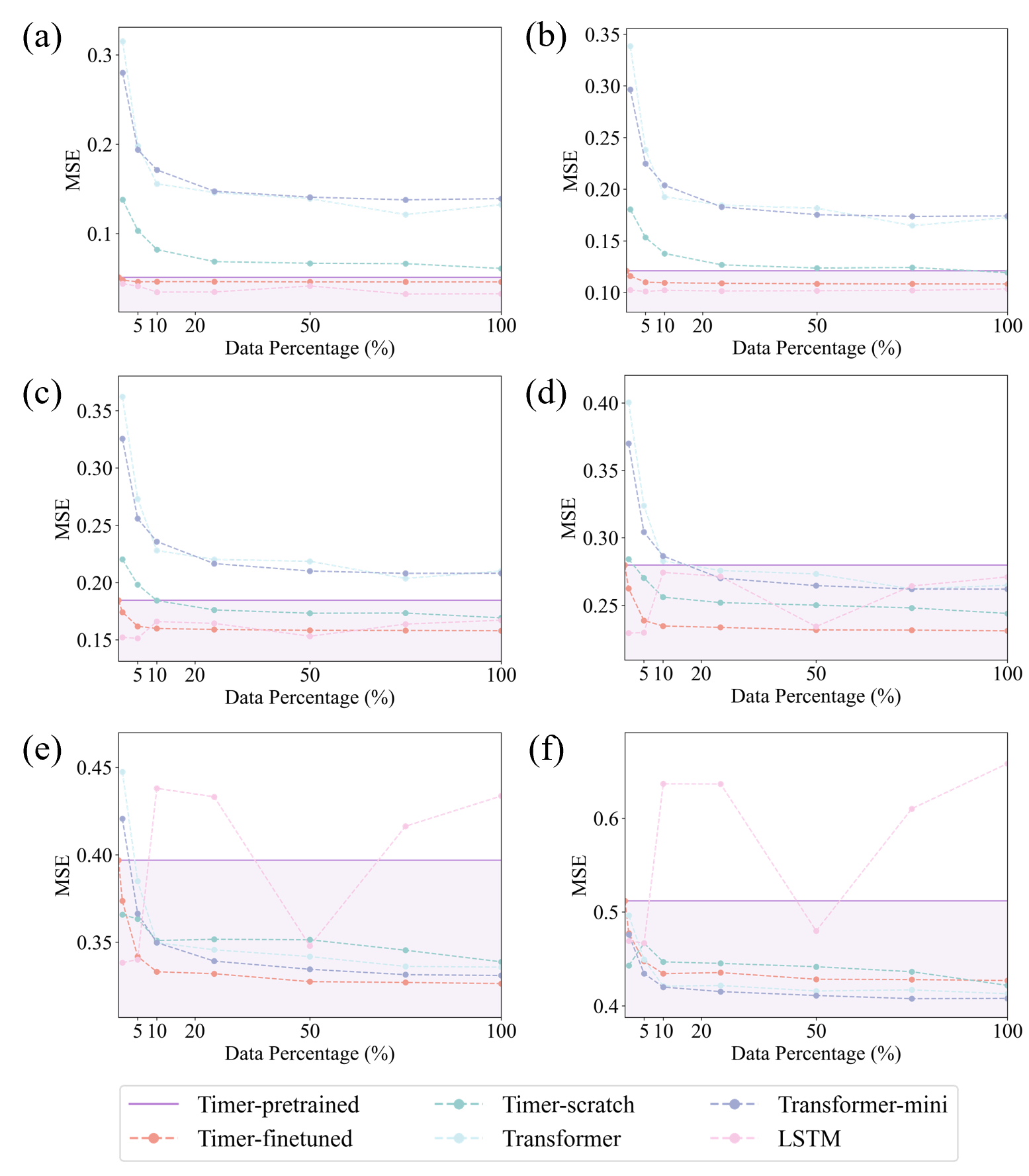}
    \caption{Variation of accuracy with data percentage. Prediction length: (a) 1. (b) 6. (c) 12. (d) 24. (e) 48. (f) 96.}
    \label{fig:fig7}
\end{figure}

To examine the influence of data volume on model accuracy, the proportion of data allocated for training was adjusted to train or fine-tune models. The test results are depicted in Figure \ref{fig:fig7}, where the accuracy of the pre-trained Timer in the zero-shot context is delineated by a purple horizontal line as a baseline. Pre-trained large time series models necessitate less data for short prediction horizons as compared to other transformer-based architectures, where accuracy remains largely stable with diminishing data volumes. Nevertheless, the benefit of pre-training wanes as the prediction horizon extends. Notably, the Timer-scratch, which has the same scale of parameters as the Transformer but lacks pre-training, does not exhibit a significant decline in accuracy. This is possibly due to its channel-independent strategy that provides stronger generalization capabilities. The LSTM still exhibits a reduced dependency on data volume, with advanced accuracy lying in short-term forecasting. However, it tend to exhibit unstable performance when tasked with long-term prediction. 

\subsection{One-turbine Fine-tuning for Whole-Plant Prediction}
Data from a single wind turbine at Plant 1 were employed to fine-tune the pre-trained large model and to train benchmarks, which were then evaluated using data from all turbines in the wind farm. This application not only investigates the influence of data volume but also evaluates the models’ generalization capabilities across various individual turbines. To mitigate the effects of randomness, the test was conducted three times, each time utilizing data from a distinct turbine. The average MSE across these three trials is detailed in Table \ref{tab:tab4}. The fine-tuned pre-trained large model demonstrated superior performance across all prediction lengths. While in the previous experiment, some models may outperform the Timer-finetuned for specific prediction lengths when the data is limited, the pre-trained large model’s generalization advantage led to its overall dominance over other models in this experiment, carrying substantial relevance for practical applications. In real-world scenarios, by collecting data from just one turbine to train a large model, its strengths in few-shot and generalization can be exploited to apply the model across other turbines throughout the plant, thereby substantially reducing the effort required for data collection and expediting the deployment of the model.

\begin{table}
    \caption{Average MSE for one-turbine training.}
    \centering
    \begin{tabular}{ccccccccc}
    \toprule
    ~ & \multicolumn{6}{c}{Prediction Length} \\
    \cmidrule(r){2-7}
    ~ & 1 & 6 & 12 & 24 & 48 & 96 \\ 
    \midrule
    Timer-finetuned & \textbf{0.0475}  & \textbf{0.1133}  & \textbf{0.1687}  & \textbf{0.2520}  & \textbf{0.3603}  & \textbf{0.4636}   \\ 
    Timer-pretrained & 0.0511  & 0.1211  & 0.1846  & 0.2799  & 0.3969  & 0.5120   \\ 
    Timer-scratch & 0.1209  & 0.1704  & 0.2133  & 0.2833  & 0.3779  & 0.4663   \\ 
    Transformer & 0.3088  & 0.3355  & 0.3589  & 0.3988  & 0.4470  & 0.5011   \\ 
    Transformer-mini & 0.2576  & 0.2832  & 0.3122  & 0.3562  & 0.4124  & 0.4761   \\ 
    LSTM & 0.1222  & 0.1542  & 0.1989  & 0.2779  & 0.3973  & 0.5410   \\ 
    \bottomrule
    \end{tabular}
    \label{tab:tab4}
\end{table}

\section{Conclusion}
\label{sec:conc}
This paper explores the application of large foundation models for time series in the prediction of wind turbine SCADA data. Two wind farms, encompassing both onshore and offshore facilities, and turbines with capacities of 1.5 MW and 4 MW, were employed to assess the predictive efficacy of the large model, Timer, and to benchmark it against alternative methodologies. The experimental results across various wind farms and data volumes suggest that the large time series model does not consistently yield a superior accuracy in data-abundant cases. In scenarios of limited data, it manifest few-shot learning advantage, although this does not notably outperform LSTM in short-term forecasting. When fine-tuned on data from an individual turbine, the large model not only demonstrates its few-shot learning capacity but also its ability to generalize across different turbines, resulting in a comprehensive accuracy enhancement compared to other models within this specific context. These findings underscore the utility of pre-trained large time series models for expedited deployment in wind farms.

In the experiments, the influences of various components within the pre-trained large model on its overall performance were examined. It is evident that the impact of pre-training on model accuracy is predominantly observed in short-term predictions. Additionally, a comparative analysis of Transformers with varying parameter counts suggests that an increase in parameters predominantly enhances short-term prediction accuracy. These observations may stem from the inherent volatility and randomness of wind resources, which introduce higher unpredictability into wind turbine monitoring data compared to other domains encountered during pre-training, with only short-term temporal patterns being effectively captured. However, in short-term prediction scenarios, such as one-step-ahead forecasting, the LSTM, which processes sequences step-by-step and has a small scale of parameters, surpasses large models that treat patches as tokens, regardless of whether data is abundant or limited, which diminishes the prominence of the advantages of pre-trained large time series model. Consequently, for time series prediction of SCADA data, further refinement of current large time series models is necessary, particularly in terms of model architecture and pre-training data selection, to broaden the applicability and enhance the performance.

\bibliographystyle{unsrt}  
\bibliography{references}

\begin{thebibliography}{10}

\bibitem{RN200}
Wei Cao, Dong Wang, Jian Li, Hao Zhou, Yitan Li, and Lei Li.
\newblock Brits: bidirectional recurrent imputation for time series, 2018.

\bibitem{RN320}
Yuwei Fan, Chenlong Feng, Rui Wu, Chao Liu, and Dongxiang Jiang.
\newblock Multiscale-attention masked autoencoder for missing data imputation
  of wind turbines.
\newblock {\em Knowledge-Based Systems}, 299:112114, 2024.

\bibitem{RN188}
Chenlong Feng, Chao Liu, and Dongxiang Jiang.
\newblock Unsupervised anomaly detection using graph neural networks integrated
  with physical-statistical feature fusion and local-global learning.
\newblock {\em Renewable Energy}, 206:309--323, 2023.

\bibitem{RN285}
Chenlong Feng, Chao Liu, and Dongxiang Jiang.
\newblock Root cause localization for wind turbines using physics guided
  multivariate graphical modeling and fault propagation analysis.
\newblock {\em Knowledge-Based Systems}, 295:111838, 2024.

\bibitem{RN296}
Yongchao Zhu, Caichao Zhu, Jianjun Tan, Yili Wang, and Jianquan Tao.
\newblock Operational state assessment of wind turbine gearbox based on long
  short-term memory networks and fuzzy synthesis.
\newblock {\em Renewable Energy}, 181:1167--1176, 2022.

\bibitem{RN290}
Shilin Sun, Yuekai Liu, Qi~Li, Tianyang Wang, and Fulei Chu.
\newblock Short-term multi-step wind power forecasting based on spatio-temporal
  correlations and transformer neural networks.
\newblock {\em Energy Conversion and Management}, 283, 2023.

\bibitem{RN313}
Jingbo Zhou, Xinjiang Lu, Yixiong Xiao, Jian Tang, Jiantao Su, Yu~Li, Ji~Liu,
  Junfu Lyu, Yanjun Ma, and Dejing Dou.
\newblock Sdwpf: A dataset for spatial dynamic wind power forecasting over a
  large turbine array.
\newblock {\em Scientific Data}, 11(1), 2024.

\bibitem{RN334}
Jeffrey~L. Elman.
\newblock Finding structure in time.
\newblock {\em Cognitive Science}, 14(2):179--211, 2010.

\bibitem{RN335}
F.~Scarselli, M.~Gori, Tsoi Ah~Chung, M.~Hagenbuchner, and G.~Monfardini.
\newblock The graph neural network model.
\newblock {\em IEEE Transactions on Neural Networks}, 20(1):61--80, 2009.

\bibitem{RN206}
Ashish Vaswani, Noam Shazeer, Niki Parmar, Jakob Uszkoreit, Llion Jones,
  Aidan~N. Gomez, Łukasz Kaiser, and Illia Polosukhin.
\newblock Attention is all you need, 2017.

\bibitem{RN336}
Jared Kaplan, Sam McCandlish, Tom Henighan, Tom~B. Brown, Benjamin Chess, Rewon
  Child, Scott Gray, Alec Radford, Jeffrey Wu, and Dario Amodei.
\newblock Scaling laws for neural language models.
\newblock {\em arXiv e-prints}, page arXiv:2001.08361, 2020.

\bibitem{RN337}
Alexander Kirillov, Eric Mintun, Nikhila Ravi, Hanzi Mao, Chloe Rolland, Laura
  Gustafson, Tete Xiao, Spencer Whitehead, Alexander~C. Berg, Wan-Yen Lo, Piotr
  Dollár, and Ross Girshick.
\newblock Segment anything.
\newblock {\em arXiv e-prints}, page arXiv:2304.02643, 2023.

\bibitem{RN216}
Aditya Ramesh, Prafulla Dhariwal, Alex Nichol, Casey Chu, and Mark Chen.
\newblock Hierarchical text-conditional image generation with clip latents.
\newblock {\em arXiv e-prints}, page arXiv:2204.06125, 2022.

\bibitem{RN211}
Tom~B. Brown, Benjamin Mann, Nick Ryder, Melanie Subbiah, Jared Kaplan,
  Prafulla Dhariwal, Arvind Neelakantan, Pranav Shyam, Girish Sastry, Amanda
  Askell, Sandhini Agarwal, Ariel Herbert-Voss, Gretchen Krueger, Tom Henighan,
  Rewon Child, Aditya Ramesh, Daniel~M. Ziegler, Jeffrey Wu, Clemens Winter,
  Christopher Hesse, Mark Chen, Eric Sigler, Mateusz Litwin, Scott Gray,
  Benjamin Chess, Jack Clark, Christopher Berner, Sam McCandlish, Alec Radford,
  Ilya Sutskever, and Dario Amodei.
\newblock Language models are few-shot learners.
\newblock {\em arXiv e-prints}, page arXiv:2005.14165, 2020.

\bibitem{RN327}
Jiexia Ye, Weiqi Zhang, Ke~Yi, Yongzi Yu, Ziyue Li, Jia Li, and Fugee Tsung.
\newblock A survey of time series foundation models: Generalizing time series
  representation with large language model.
\newblock {\em arXiv e-prints}, page arXiv:2405.02358, 2024.

\bibitem{RN291}
Nate Gruver, Marc Finzi, Shikai Qiu, and Andrew~Gordon Wilson.
\newblock Large language models are zero-shot time series forecasters.
\newblock {\em arXiv e-prints}, page arXiv:2310.07820, 2023.

\bibitem{RN292}
Ming Jin, Shiyu Wang, Lintao Ma, Zhixuan Chu, James~Y. Zhang, Xiaoming Shi,
  Pin-Yu Chen, Yuxuan Liang, Yuan-Fang Li, Shirui Pan, and Qingsong Wen.
\newblock Time-llm: Time series forecasting by reprogramming large language
  models.
\newblock {\em arXiv e-prints}, page arXiv:2310.01728, 2023.

\bibitem{RN328}
Mingtian Tan, Mike~A. Merrill, Vinayak Gupta, Tim Althoff, and Thomas
  Hartvigsen.
\newblock Are language models actually useful for time series forecasting?
\newblock {\em arXiv e-prints}, page arXiv:2406.16964, 2024.

\bibitem{RN314}
Yong Liu, Haoran Zhang, Chenyu Li, Xiangdong Huang, Jianmin Wang, and Mingsheng
  Long.
\newblock Timer: Generative pre-trained transformers are large time series
  models.
\newblock {\em arXiv e-prints}, page arXiv:2402.02368, 2024.

\bibitem{RN317}
Yuqi Nie, Nam~H. Nguyen, Phanwadee Sinthong, and Jayant Kalagnanam.
\newblock A time series is worth 64 words: Long-term forecasting with
  transformers.
\newblock {\em arXiv e-prints}, page arXiv:2211.14730, 2022.

\bibitem{RN342}
Lu~Han, Han-Jia Ye, and De-Chuan Zhan.
\newblock The capacity and robustness trade-off: Revisiting the channel
  independent strategy for multivariate time series forecasting.
\newblock {\em arXiv e-prints}, page arXiv:2304.05206, 2023.

\bibitem{RN343}
Thomas Wang, Adam Roberts, Daniel Hesslow, Teven Le~Scao, Hyung~Won Chung,
  Iz~Beltagy, Julien Launay, and Colin Raffel.
\newblock What language model architecture and pretraining objective work best
  for zero-shot generalization?
\newblock {\em arXiv e-prints}, page arXiv:2204.05832, 2022.

\bibitem{RN344}
Damai Dai, Yutao Sun, Li~Dong, Yaru Hao, Shuming Ma, Zhifang Sui, and Furu Wei.
\newblock Why can gpt learn in-context? language models implicitly perform
  gradient descent as meta-optimizers.
\newblock {\em arXiv e-prints}, page arXiv:2212.10559, 2022.

\end{thebibliography}

\end{document}